\def\year{2020}\relax
\documentclass[letterpaper]{article} 
\usepackage{aaai20}  
\usepackage{times}  
\usepackage{helvet} 
\usepackage{courier}  
\usepackage[hyphens]{url}  
\usepackage{graphicx} 
\usepackage{graphicx}  
\usepackage{footnote}
\usepackage{makecell}
\usepackage{booktabs}

\usepackage{amsmath}
\DeclareMathOperator*{\argmax}{arg\,max}

\newcommand{\newcite}[1]{\citeauthor{#1} \shortcite{#1}}

\nocopyright
\title{A Universal Parent Model for Low-Resource Neural Machine Translation Transfer }
\author{Mozhdeh Gheini, Jonathan May\\ \Large USC Information Sciences Institute\\ \Large \texttt{\{gheini, jonmay\}@isi.edu}}
 
\begin{document}

\maketitle

\begin{abstract}
Transfer learning from a high-resource language pair `parent' has been proven to be an effective way to improve neural machine translation quality for low-resource language pairs `children.' However, previous approaches build a custom parent model or at least update an existing parent model's vocabulary for each child language pair they wish to train, in an effort to align parent and child vocabularies. This is not a practical solution. It is wasteful to devote the majority of training time for new language pairs to optimizing parameters on an unrelated data set. Further, this overhead reduces the utility of neural machine translation for deployment in humanitarian assistance scenarios, where extra time to deploy a new language pair can mean the difference between life and death. In this work, we present a `universal' pre-trained neural parent model with constant vocabulary that can be used as a starting point for training practically any new low-resource language to a fixed target language. We demonstrate that our approach, which leverages orthography unification and a broad-coverage approach to subword identification, generalizes well to several languages from a variety of families, and that translation systems built with our approach can be built more quickly than competing methods and with better quality as well. 
\end{abstract}

\section{Introduction}
\label{sec:intro}





As has been previously shown \cite{zoph2016transfer} and as is confirmed in Table~\ref{tab:res}, for low-resource language pairs, \textit{transfer learning} from a much larger `parent' language pair leads to higher quality results, particularly when the target language is the same in both cases. However, it is not entirely clear a) why this is, or b) what a `good' parent should be. More practically, low-resource machine translation (MT) systems often arise out of necessity and with immediacy. The sudden discovery that there is a need to translate a language for which few resources have been previously collected suggests an emergency situation, with lives potentially on the line \cite{crisis-mt-developing-a-cookbook-for-mt-in-crisis-situations}. This makes the approach of building a custom parent model, for each new child language, as was done previously \cite{zoph2016transfer,nguyen2017transfer,kocmi2018trivial}, infeasible. Rather than spend days or weeks to build a new parent model before transferring to the intended language pair and fine-tuning model weights, we would greatly prefer to have a single model at the ready, capable of being quickly fine-tuned toward any new language pair and deployed within hours.

Our desire for a universal parent neural translation model is related to our desire to understand why, in low-resource scenarios, it is still often the case that non-neural statistical translation models (SMT) outperform neural counterparts \cite{Koehn2017SixCN}. We suggest that there is a key difference in the \textit{untrained} models of each paradigm: Even when untrained, SMT is nonetheless a \textit{translation model}, biased in terms of feature choice, search heuristic, and objective specifically for translating between human languages. On the other hand, neural models, whether Transformer \cite{vaswani2017attention}, recurrent \cite{sutskever2014sequence,bahdanau2014neural,luong-pham-manning:2015:EMNLP}, or convolutional \cite{gehring2017convolutional}, are, before training, simply  \textit{transduction} models, designed to represent conditionally weighted sequence pairs, but no more designed to translate languages than to paraphrase, engage in dialogue, or answer questions. We theorize that a universal pre-trained parent neural model can capture some of the `baked-in' universal properties of translation analogous to those engineered into SMT models and will thus be more amenable to transfer and subsequent fine-tuning on smaller new language data sets, since they will not have to re-learn the fundamentals of translation from scratch.

In this work, we propose such a \textit{universal} parent model that additionally contains a pre-learned and fixed vocabulary for translating from any language into English. While potentially useful on its own and in this sense closely related to other multilingual systems \cite{johnson2017google}, the true value of this model is its enabling of rapid generation of neural translation models for new languages \textit{without} requiring any retraining of the parent or updating its vocabulary. To do this we:
\begin{itemize}
    \item Analyze the potential reasons that transfer learning helps neural machine translation (NMT) and determine that the procedure does in fact ensure the models are `ready-to-translate,' and not simply improved by language model enhancements or word borrowing; 
    \item Analyze lexical issues that can arise from naively building a subword tokenization model such as a byte-pair encoding (BPE) \cite{sennrich2015neural} for a parent language before a child language is known and introduce a many-language BPE that avoids this issue; and
    \item Consequently, introduce a wide-coverage universal training regimen that makes use of universal romanization \cite{hermjakob-etal-2018-box} to avoid orthography issues by incorporating a universal subword vocabulary \cite{wu2016google}, thus yielding a pre-trained parent that can be used out-of-the-box with any new language pair.  
\end{itemize}


\begin{table*}[h]
    \centering
    \begin{tabular}{|l|c|c|c|c|c|}
        \cline{2-6}
        \multicolumn{1}{c|}{} & \textbf{Ro-En} & \textbf{Hi-En} & \textbf{Lt-En} & \textbf{Fi-En} & \textbf{Et-En} \\
        \hline
        Corpus & WMT16 & IITB & \multicolumn{3}{c|}{Europarl} \\
        \hline
        Subset Size Used (Sentences) & 60k & 35k & 43k & 20k & 10k \\
        \hline
        Baseline & 24.93 & 2.39 & 10.91 & 4.19 & 2.91 \\
        \hline
        French Parent & 26.83 & 6.77 & 12.56 & 7.60 & 5.71 \\
        \hline
        \makecell[cl]{French Parent with \\ Universal Subwords} & 27.10 & 7.18 & 13.24 & 7.54 & 5.92 \\
        \hline
        Universal Parent & \textbf{\underline{27.35}} & \textbf{\underline{9.10}} & \textbf{\underline{13.82}} & \textbf{\underline{8.52}} & \textbf{\underline{6.49}} \\
        \hline
    \end{tabular}
    \caption{Training corpus information and test scores over 5 child languages. To simulate a low-resource setting, we sample a subset of the corpus available for each language. Consistent improvements can be observed along columns, across languages. Using a universal parent, which is able to train a universal vocabulary well, yields the best results.}
    \label{tab:res}
\end{table*}

\section{Background}
\label{sec:background}

\subsection{Neural Machine Translation}
\begin{figure}[h]
    \centering
    \includegraphics[width=0.8\columnwidth]{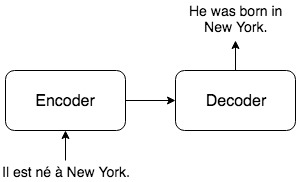}
    \caption{Schematic diagram of the encoder-decoder architecture. French `Il est ne a New York.' is encoded to be later decoded as English `He was born in New York.'.}
    \label{fig:encdec}
\end{figure}

Machine translation is the task of generating the corresponding target language sequence $Y = (y_1, y_2, ..., y_m)$ given the source language sequence $X = (x_1, x_2, ..., x_n)$. This is achieved by maximizing the conditional probability of $p(Y|X)$. So $$Y = \argmax_Yp(Y|X).$$
Neural machine translation typically achieves this by creating a representation of $X$ which is used to condition the sequential generation of the words $y_i$ of $Y$, along with $y_1, \ldots, y_{i-1}$. This is known as the \textit{encoder-decoder} paradigm  \cite{cho2014learning} and is typically learned via minimization of $-\log p(Y|X)$. We provide an illustration in Figure~\ref{fig:encdec}. The choice of architecture for the encoder and the decoder themselves can vary; typical approaches used recently include Transformers \cite{vaswani2017attention} (used in this work), RNNs \cite{sutskever2014sequence}, or CNNs \cite{gehring2017convolutional}. The encoder-decoder architecture can also be further improved by using an attention mechanism between the decoder and the encoder \cite{bahdanau2014neural,luong-pham-manning:2015:EMNLP}; indeed, this constitutes the majority of the design behind Transformer \cite{vaswani2017attention}.

\subsection{Transfer Learning}
Transfer learning is based on the intuition that more often than not we don't actually need to train a model from scratch; in fact, we usually don't have even close to enough data for that. Instead, we can benefit from the knowledge learned during training some other related task to train on our current task.

For a formal definition, we can refer to \newcite{pan2009survey}. \citeauthor{pan2009survey} first define a domain ($\mathcal{D}$) as a tuple of a feature space ($\mathcal{X}$) and a probability distribution $P(X=\boldsymbol{x}\in \mathcal{X})$ over it. Then, given a domain, they define a task ($\mathcal{T}$) as a tuple of a label space ($\mathcal{Y}$) and a predictive function $f:\mathcal{X}\rightarrow\mathcal{Y}$, which can be learned from training data. Given these two definitions, they define transfer learning as:

\begin{quote}
Given a source domain $\mathcal{D}_S$ and learning task $\mathcal{T}_S$, a target domain $\mathcal{D}_T$ and learning task $\mathcal{T}_T$, transfer learning aims to help improve the learning of the target predictive function $f_T$ in $\mathcal{T}_T$ using the knowledge in
$\mathcal{D}_S$ and $\mathcal{T}_S$, where $\mathcal{D}_S\neq\mathcal{D}_T$, or $\mathcal{T}_S\neq\mathcal{T}_T$.
\end{quote}

In practice, most of the time transfer learning is embodied by means of fine-tuning, where a new model is initialized with the parameters of another trained model and its training starts from there.

\subsection{Subword Vocabulary}
Translation systems, whether statistical or neural, work with fixed-sized vocabularies, usually of at most a few thousand types. Given that translation, like many other NLP tasks, is open-vocabulary, this leaves all models with the problem of translating rare and unseen words, referred to as out-of-vocabulary (OOV) words.

Following the inspiration from human translation where never-before-seen words are often translated by translating before-seen smaller units, such as morphemes, many solutions proposed for machine translation of unknown words also involve operating at the subword level rather than word level.

If word-level translation is at one extreme end, on the other end we have character-level translation \cite{tiedemann2009character,neubig2012machine,ling2015character,luong2016achieving,costa2016character,chung2016character,lee2017fully}. Falling somewhere in the middle, an appropriate \textit{word segmentation} method should divide words into maximally reusable but also maximally semantically meaningful sub-units. Inspired by the BPE data compression algorithm \cite{gage1994new}, \newcite{sennrich2015neural} propose a word segmentation algorithm where an initially character-segmented text is repeatedly scanned, and the most frequently occurring token pairs are iteratively merged. The number of times this merge operation is done is an algorithm hyperparameter. In the end, frequent character sequences are merged into and represented with single symbols, and these symbols form our subword vocabulary. We provide a toy example detailed in Table~\ref{tab:bpeexmpl}. It lists the first five merge operations given the dictionary \{`the', `mother', `father', `fewer'\}, where we assume `the' occurs 3 times, `mother' occurs 1 time, `father' occurs 1 time, and `fewer' occurs 2 times in the corpus.

\begin{table}[h]
    \centering
    \begin{tabular}{c|c|c}
        \toprule
        \multicolumn{3}{c}{Dictionary and Counts} \\
        \multicolumn{3}{c}{\{`the\_': 3, `mother\_': 1, `father\_': 1, `fewer\_': 2\}} \\
        \midrule
        Pair to Merge & Pair Frequency & New Symbol Created  \\
        \midrule
        t h & 5 & th \\
        th e & 5 & the \\
        e r & 4 & er \\
        er \_ & 4 & er\_ \\
        the \_ & 3 & the\_ \\
        \bottomrule
    \end{tabular}
    \caption{BPE merge operations for the given dictionary. The algorithm first marks the ends of the words by appending a symbol to them. Here, we use \_ as the end-of-word symbol. Then as shown, it iteratively merges the most frequent symbol pair into one symbol. For instance, at the first step, the character bigram `t h' gets merged into `th' as the most frequent pair.}
    \label{tab:bpeexmpl}
\end{table}

\section{Transfer Learning For Low-Resource MT}
\label{sec:method}

The basic idea in transfer learning for machine translation is to, instead of initializing randomly, initialize a child model with the trained parameters of a parent model and then fine-tune it on the child language pair \cite{zoph2016transfer}. Although straightforward for most parameters, when it comes to transferring embeddings, the weights corresponding to the model's vocabulary, it is not necessarily clear how to do this assignment. After all, parent and child models are specifically different in the language they are translating, so under normal circumstances their vocabularies are in fact quite different. While one could theoretically use a sufficiently large parallel lexicon to reassign initial embeddings,  these lexicons are quite difficult to obtain and have various problems including polysemy and untranslatable concepts. In practice,  an arbitrary or frequency-based reassignment is done \cite{zoph2016transfer}, but ideally we would like to \textit{not reassign at all}. In order to avoid reassignment, then, the vocabularies of parent and child must be the same. While this can be done for character models, the information content in these embeddings is questionable and in practice such models have not been shown to be practical in speed or quality. We would like the vocabulary to contain more semantic heft than characters can provide.
The general workaround, however, is to \textit{bias} the parent vocabulary, e.g., by using vocabulary from concatenated parent and child corpora \cite{zoph2016transfer}. When time is no object and a new parent model can be constructed for each child model, this is a feasible approach, but in the face of a true surprise language this is, of course, an unrealistic assumption.

Another approach has been to update the vocabulary of the parent model at the time of fine-tuning by taking the vocabulary obtained from the child corpus into account. This has been done either by simply adding the child vocabulary to the parent's \cite{neubig2018rapid,lakew2018transfer} or by matching parent and child embedding spaces by learning a cross-lingual projection from child embedding space to parent embedding space \cite{kim2019effective}. Nevertheless, the former causes a significant increase in the model vocabulary size, and the latter, again, considerably delays the fine-tuning process. Specifically, in the latter, we need to learn monolingual child embeddings (assuming we have a large enough monolingual corpus) and cross-lingual linear mapping between the child and parent spaces.

In this work, we show that a universal vocabulary, where subwords \cite{wu2016google} are obtained from as many languages as available, provides both an efficient vocabulary in the face of a new language and a fast way to have a \textit{ready-to-go model with fixed-sized vocabulary} to fine-tune on any child. This universal vocabulary does not need to necessarily have seen vocabulary from either the child or parent (similar to \cite{johnson2017google}). In fact, as we explain in Section~\ref{sec:exp}, we only experiment on child languages \textit{not} included in obtaining the universal vocabulary. To make sure that this universal vocabulary works for all languages we unite the orthographies by using the universal romanizer \textit{uroman} \cite{hermjakob-etal-2018-box}.

To demonstrate the success of universal vocabulary in transfer learning, we run several sets of experiments: For the baseline, we simply train a model from scratch on our low-resource language pair. 
A French parent model trained on French vocabulary replicates the conditions in most previous work, e.g. \cite{zoph2016transfer}.
Our main contribution, a nineteen-language multilingual parent model and universal vocabulary is shown to be more powerful across multiple child language pairs.
Finally, several ablative conditions tease apart the impact of language model, subword model, and source contribution.

\section{Experimental Setup}
\label{sec:exp}

\subsection{Data}
We obtain data for our parent and child models from different sources. For the universal parent model, we use the data from 19 language pairs (with all having English as their target side and the sources being Akan, Amharic, Arabic, Bengali, Persian, Hausa, Hungarian, Indonesian, Russian, Somali, Spanish, Swahili, Tamil, Tagalog, Turkish, Uzbek, Vietnamese, Wolof, and Yoruba), prepared for the DARPA-LORELEI program \cite{Christianson2018}. The size of each corpus ranges from ${\sim}\text{27k}$ sentences (Indonesian) to ${\sim}\text{387k}$ sentences (Spanish). 
When put together, this gives us a parallel multilingual corpus of around 2 million sentences. For the French parent, we use the Giga French-English Corpus \cite{callisonburch-EtAl:2009:WMT}.\footnote{\url{http://www.statmt.org/wmt15/translation-task.html}} To make the French comparable in size with the universal parent we only use 2 million sentences from the 22 million sentences available in the Giga corpus.

We experiment with and report results on 5 child languages: Romanian, Hindi, Lithuanian, Finnish, and Estonian. For each one, we use one of WMT16,\footnote{\url{http://www.statmt.org/wmt16/translation-task.html}} IITB \cite{kunchukuttan2017iit},
 and Europarl\footnote{\url{http://www.statmt.org/europarl}} corpora. The corpora and the amount of data we use for each are detailed in Table~\ref{tab:res}. 

All data is romanized using uroman \cite{hermjakob-etal-2018-box} and then used as input to the translation system. Universal subwords are formed by the vocabularies from the concatenation of the 19 languages and their English translations. Ablative experiments using alternate subword vocabularies in Section~\ref{sec:disc}, use parallel data with the source as indicated. 

\subsection{NMT system}
We use the tensor2tensor \cite{tensor2tensor} implementation of the Transformer \cite{vaswani2017attention} to train our models. We use a vocabulary size of 8k subword pieces when training baseline models, and a vocabulary size of 16k subword pieces when training and transferring our parent models. Other than that, we use the default hyperparameters and train the transformer base model as described in \newcite{vaswani2017attention}, maintaining a single merged embedding vocabulary and parameter set for both source and target languages.

\section{Results}
\label{sec:res}

We report our core results in Table~\ref{tab:res}. While simply transferring from the French parent model with French vocabulary using \newcite{zoph2016transfer}'s approach is better than training from scratch, with no parent model (row 2), there are benefits to be gained from the universal vocabulary. Specifically, even though French itself is not included in the 19 languages used to obtain the universal vocabulary, French serves as a better parent if universal vocabulary is used (row 3). As expected, the best results across all five language pairs are obtained if we use the universal parent data as well (row 4). Overall we are able to gain as much as a 6.71 BLEU increase over the baseline (row 1) and 2.33 BLEU increase over vanilla transfer learning (row 2) using the universal parent.

\section{Discussion}
\label{sec:disc}

In this section we try to answer three questions: 1) Do common subword pieces help translation, or are improvements by both multilingual training and transfer learning just an effect of an improved target language model thanks to more data? 2) If subword pieces do help, how do they help? and finally, 
3) Can large amounts of monolingual data win over universality? In other words, does one need to bother training a multilingual model with shared vocabulary if they have a larger monolingual parallel corpus that may actually benefit them more?

To answer these questions, we first focus on the Ro-En data and carry out ablative experiments, the results of which we report in Table~\ref{tab:disc}. In the `model' column of the table, whenever data is concatenated to train a multilingual model, we use +, and whenever we transfer a model and further fine-tune it, we use $\rightarrow$.

\begin{table}[h]
    \centering
    \begin{tabular}{|c|c|c|c|}
        \hline
        \# & Model & \makecell{Subwords \\ Used} & \makecell{Ro Test \\ BLEU} \\
        \hline
        1 & Ro & Ro & 24.93 \\
        \hline
        2 & Fr+Ro & Fr+Ro & 25.47 \\
        \hline
        3 & CJKFr+Ro & CJKFr+Ro & 25.01 \\
        \hline
        4 & Fr $\rightarrow$ Ro & Fr & 26.83 \\
        \hline
        5 & Fr $\rightarrow$ Ro & Univ & 27.10 \\
        \hline
        6 & Univ $\rightarrow$ Ro & Univ & 27.35 \\
        \hline
        7 & Fr $\rightarrow$ Ro & Fr+Ro & 28.09 \\
        \hline
        8 & Fr+Ro $\rightarrow$ Ro & Fr+Ro & 27.77 \\
        \hline
        9 & CJKFr+Ro $\rightarrow$ Ro & CJKFr+Ro & 27.42 \\
        \hline
    \end{tabular}
    \caption{Romanian analysis experiments. `Univ' refers to the universal parent discussed before. `CJKFr' refers to French replaced with non-latin characters to avoid sharing any subword information}
    \label{tab:disc}
\end{table}

To answer the first question, that of if common subword pieces help or if improvements are mainly due to an improved language model, we replace all non-punctuation characters from the source side of the French-English corpus with non-Latin characters that are thus guaranteed to not share any subsequences with the Romanian or English parts of the vocabulary; we label this `CJKFr.'\footnote{Our implementation randomly chooses characters from the Unicode CJK table to replace the Latin characters.} {CJKFr} is virtually French, only with a new character set. If the improved results were just due to an improved language model, {CJKFr} would have been able to perform as well as French, since the data on the target side is the same in both cases. However, we can see that both in multilingual training (rows 2 and 3) and transfer learning (rows 8 and 9), {CJKFr} is not able to perform as well as French.

We also report results of when we incorporate child vocabulary knowledge in the parent by training a vocabulary on concatenated corpora (row 7), and indeed it yields the best results. So with enough time at hand, it is best to train a custom parent with knowledge of the child vocabulary. However, in the face of an emergency, a universal parent is able to give comparable results in much less time. In our experiments, while the child models need 20k-50k steps to converge, the parent needs over 120k steps. The converged parent model gets BLEU of 28.86, but it only scores 23.40 and 26.67 after training for 20k and 50k steps, respectively. 

To determine how common subwords can help, we define the `over-segmentation rate` heuristic. We say a word is \textit{over-segmented} by a vocabulary if the number of the segments is more than half of the length of the unsegmented word, meaning there is at least one single character in the resulting segmentation. We ideally want words to not be over-split. When the subwords from French are used (row 4) we get an over-segmentation rate of 0.04 on Romanian data vs an over-segmentation rate of 0.02 when we use subwords from the universal vocabulary (rows 5 and 6). For Finnish, which is not as close to French, the over-segmentation rates are 0.01 (univ.) vs. 0.1 (French). The over-segmentation rates for all languages we reported results on in Table~\ref{tab:res} are provided in Table~\ref{tab:segrate}.

\begin{table}[h]
    \centering
    \begin{tabular}{|r|c|c|}
        \cline{2-3}
        \multicolumn{1}{c|}{} & Univ. Vocab. OsR & French Vocab. OsR \\
        \hline
        \textbf{Ro} & 0.02 & 0.04 \\
        \hline
        \textbf{Hi} & 0.03 & 0.14 \\
        \hline
        \textbf{Lt} & 0.02 & 0.09 \\
        \hline
        \textbf{Fi} & 0.01 & 0.11 \\
        \hline
        \textbf{Et} & 0.01 & 0.09 \\
        \hline
    \end{tabular}
    \caption{Over-segmentation rates (OsR) for all languages we experiment with using a monolingual French subword vocabulary vs. a multilingual universal subword vocabulary. The lower the over-segmentation rate, the fewer words were over-segmented by the given subword vocabulary. In all case, using the universal vocabulary leaves us with lower OsR.}
    \label{tab:segrate}
\end{table}

Perhaps some examples from Finnish, a language with high morphological inflection, can illustrate this better. The word `liikenneruuhkien' in Finnish means `traffic jam' and is constructed by concatenating `liikenne' (transportation, traffic) and `ruuhkien' (congestion). When segmented by French subwords, it turns into `li ik enn er u u h ki en\_' and when segmented by universal subwords, it turns into `lii ken ner uu hki en\_' (an \_ indicates the end of the word). As another example, take `polttoainevarannot', which means `fuel reserves' in Finnish and is made from smaller words `polttoaine' (fuel) and `varannot' (reserves). When segmented by French subwords, it turns into `pol tt oa ine va ran not\_' and when segmented by universal subwords, it turns into `pol tto aine vara nno t\_'. In both cases the French subwords break the words into more segments than the universal subwords do; in the first case even down to single characters. The examples in Table~\ref{tab:exmpl} shows how this can impact translation output.

We can gain even more insight into how these shared subwords get reused by looking at the segmentations of the same word in different languages when we feed them through the universal vocabulary romanized in several languages. Table~\ref{tab:philippines} show segmentations of the word `Philippines' in 4 languages included in the universal parent (namely Russian, Somali, Hausa, and Yoruba). We can observe that subwords `Fili' and `ppi' are reused among these languages. More importantly, again an example from Finnish, when the universal vocabulary faces the new word `Filippiinit' (`Philippines' in Finnish), it breaks it down to  `Fili ppi ini t\_', reusing the same subwords it has learned from and seen before in the parent model languages. This is while, the French vocabulary breaks `Filippiinit' down to 'Fil ip pi ini t\_'; not only does its segmentation have more subwords than the universal vocabulary's segmentation (5 vs. 4), it's also unlikely those subwords carry much semantic weight given that `Philippines' in French is `Philippines' and doesn't share any subwords included in 'Fil ip pi ini t\_'.

\begin{table}[h]
    \centering
    \begin{tabular}{c|c}
        \toprule
        \makecell{`Philippines' romanized \\ in Language X} & \makecell{Universal Vocab. \\ Segmentation} \\
        \midrule
        Filippiny & Fili ppi ny\_ \\
        Filibiin & Fili bi in\_ \\
        Filifin & Fili fin\_ \\
        Filippin & Fili ppi n\_ \\
        \bottomrule
    \end{tabular}
    \caption{`Philippines' in 4 languages romanized and their universal vocabulary segmentations.}
    \label{tab:philippines}
\end{table}

\begin{table*}[h]
    \centering
    \begin{tabular}{r|p{5in}}
        \toprule
        Source & Yksi haenen tavoitteistaan olikin liikenneruuhkien vaelttaeminen. \\
        French vocab. segmentation & Y ks i\_ ha ene n\_ ta voit te ist a an\_ oli kin \_ li ik enn er u u h ki en\_ va el tta em ine n\_  .\_ \\
        Universal vocab. segmentation & Y ksi \_ ha ene n\_ tav oit te ista an\_ oli kin\_ lii ken ner uu hki en\_ va elt tae mine n\_  .\_ \\
        \hline
        Reference &  One of his goals was to  avoid traffic jam. \\
        \hline
        French parent output &  One of his objectives was to avoid mobility. \\
        \hline
        Universal parent output &  One of his goals was to avoid traffic jams. \\
        \midrule
        \midrule
        Source & Myoes polttoainevarannot ovat ehtymaessae. \\
        French vocab. segmentation & My oe s\_ pol tt oa ine va ran not\_ ov at\_ e ht ym aes sa e\_  .\_ \\
        Universal vocab. segmentation & My oes \_ pol tto aine vara nno t\_ ovat\_ eh ty mae ssa e\_  .\_ \\
        \hline
        Reference &  Also fuel reserves are running short. \\
        \hline
        French parent output &  This too is the case. \\
        \hline
        Universal parent output &  fuel reserves are also under pressure. \\
        \bottomrule
    \end{tabular}
    \caption{Example sentences from the Finnish test set. The system transferred from French parent with French subwords completely misses `congestion' and `fuel reserves' due to over-segmenting `ruuhkien' and `polttoainevarannot', respectively.}
    \label{tab:exmpl}
\end{table*}

To answer the third question, regarding whether the same benefits can be gained by simply using a large enough monolingual corpus, we train yet another monolingual parent, using 11 million parallel sentences from the Giga French-English corpus, mentioned earlier in Section~\ref{sec:exp}. We also transfer this parent to our five child languages from Table~\ref{tab:res} in the same manner as we did our smaller French parent, which was trained on only 2 million sentences from the Giga corpus, and compare its results with the results from our universal parent trained on ${\sim}\text{2}$ million sentence multilingual parallel corpus. Furthermore, to have a totally fair comparison, we also prepare an ${\sim}\text{11}$ million sentence multilingual parallel corpus by concatenating our present multilingual corpus (${\sim}\text{2}$m sentences), TED multilingual corpus (${\sim}\text{5}$m sentences) \cite{qi2018and}, and Europarl German and Spanish corpora (${\sim}\text{5}$m sentences). Once more, we make sure we strip all corpora from languages we want to test on to make sure they are completely new to the eye of the big universal parent model and its subword vocabulary. We report the results of these experiments in Table~\ref{tab:res2}. In the table, we use `small' to indicate we are talking about a parent trained on a 2 million sentence corpus and `big' to indicate we are talking about a parent trained on an 11 million sentence corpus.

It can be observed that except for Romanian, which is close to French, even the \textit{small} universal parent can beat the \textit{big} monolingual French parent. Also, per our intuition, the \textit{big} universal parent is able to come on top in all cases.

\begin{table*}[h]
    \centering
    \begin{tabular}{|l|c|c|c|c|c|}
        \cline{2-6}
        \multicolumn{1}{c|}{} & \textbf{Ro-En} & \textbf{Hi-En} & \textbf{Lt-En} & \textbf{Fi-En} & \textbf{Et-En} \\
        \hline
        Big French Parent & 27.52 & 7.10 & 13.81 & 7.96 & 5.83 \\
        \hline
        Small Universal Parent & 27.35 & 9.10 & 13.82 & 8.52 & 6.49 \\
        \hline
        Big Universal Parent  & \textbf{\underline{28.01}} & \textbf{\underline{9.20}} & \textbf{\underline{14.73}} & \textbf{\underline{9.07}} & \textbf{\underline{7.75}} \\
        \hline
    \end{tabular}
    \caption{Test scores over 5 child languages from different-sized parents. Per scores above, the gains from multilinguality and shared vocabulary cannot simply be made up for by using a large amount of monolingual data.}
    \label{tab:res2}
\end{table*}

\section{Related Work}
\label{sec:rel}

Over the past years, to improve NMT and make it at par with SMT in the low resource case, three solutions have been worked on to battle the small size of parallel data available: multilingual training \cite{johnson2017google}, transfer learning \cite{zoph2016transfer}, and back-translation \cite{sennrich2015improving}.

Close to the focus of this work, transfer learning and enabling universal machine translation, \newcite{nguyen2017transfer} and \newcite{kocmi2018trivial} separately extended \newcite{zoph2016transfer}'s approach. However, they both make their BPE/subword vocabulary using the union of both parent and child corpora.

Perhaps the most in line with our efforts to apply multilinguality to achieve a transfer-ready model, are \newcite{neubig2018rapid} and \newcite{kim2019effective}. \citeauthor{neubig2018rapid} train a massively multilingual parent model on top of the TED corpus \cite{qi2018and}, and transfer it to child languages under warm start (child language included at the time of the training the parent) and cold start (child language introduced only at the time of fin-tuning) scenarios. They also propose similar language regularization, where at the time of fine-tuning, they append data from a similar language to the child data to avoid over-fitting to the child language, as it's often a low-resource language. The main difference between our method and \citeauthor{neubig2018rapid}'s, is that while we create our vocabulary once, and only once jointly on the multilingual parent corpus altogether and use it for any given new language, they create their vocabulary separately on each language in the multilingual corpus, and later further update it with vocabulary created for each new child language. Not only does this result in a huge vocabulary (above 300k), it also leaves us with the need to obtain a vocabulary for each new language. However, in our case, we have a \textit{constant, fixed-sized} vocabulary that accompanies our ready-to-go parent model. \citeauthor{kim2019effective}, on the other hand, approach the vocabulary mismatch between the parent and child by learning a cross-lingual mapping between their embedding spaces. This, however, needs monolingual embedding training and cross-lingual projection training for each new child language which is very time-intensive.


\newcite{lakew2018transfer} also proposed a method based on \newcite{zoph2016transfer}'s approach. They use a dynamic vocabulary (similar to \newcite{neubig2018rapid}), updated for each new language pair. This increases the size of their model with each new child language and requires a preprocessing phase for each new transfer session to update the vocabulary.

In order to make universal machine translation possible, \newcite{gu2018universal} extend multilingual NMT by suggesting a universal representation technique, where all tokens in all languages are represented as a mixture of a basis for the universal token space, which in their case is English embeddings. However, they first need large amounts of monolingual data to train monolingual embeddings and also have a more time-consuming pipeline in the face of a new language.

\section{Conclusion}

In this work, we proposed a universal training method using a universal parent model and vocabulary that is able to be trained reasonably fast and perform comparably well when faced with an arbitrary new language pair. The main advantage of our method is that  it provides a multilingual parent model accompanied with a fixed-sized vocabulary, which can be used for fine-tuning on any arbitrary new language with no need of any further intermediary action involving the new language data, such as vocabulary creation or even the training of a new custom parent model. 


\section*{Acknowledgments}

The authors would like to thank David Chiang for his insightful comments and constructive discussions on this work.


\bibliography{aaai20}
\bibliographystyle{aaai}

\end{document}